\documentclass[letterpaper, 10 pt, conference]{ieeeconf}  % Comment this line out if you need a4paper

\IEEEoverridecommandlockouts                              % This command is only needed if 
\usepackage{graphicx} % for pdf, bitmapped graphics files

\usepackage{algorithm}
\usepackage{algorithmic}
\usepackage{cprotect}
\usepackage[hidelinks]{hyperref}
\usepackage{makecell}
\usepackage{multirow}
\usepackage{subcaption}
\usepackage{textcomp}

\title{\LARGE \bf
Exploring Adversarial Obstacle Attacks in\\
Search-based Path Planning for Autonomous Mobile Robots
}

\author{Adrian Szvoren$^{1}$, Jianwei Liu$^{1}$, Dimitrios Kanoulas$^{1\ast}$, Nilufer Tuptuk$^{2\ast}$%
\thanks{$^{1}$Adrian Szvoren, Jianwei Liu, and Dimitrios Kanoulas are with the Department of Computer Science, University College London, Gower Street, WC1E 6BT, London, UK. \{adrian.szvoren.23, jianwei.liu.21, d.kanoulas\}@ucl.ac.uk}%
\thanks{$^{2}$Nilufer Tuptuk is with the Department of Security and Crime Science, University College London, Gower Street, WC1E 6BT, London, UK. \{n.tuptuk\}@ucl.ac.uk}%
\thanks{$\ast$ Equal contribution.}%
\thanks{This work was supported by the UKRI FLF [MR/V025333/1] (RoboHike) and the CDT in Cybersecurity [EP/S022503/1]. For the purpose of Open Access, the author has applied a CC BY public copyright license to any Author Accepted Manuscript version arising from this submission.}%
}

\begin{document}

\maketitle
\thispagestyle{empty}
\pagestyle{empty}

%%%%%%%%%%%%%%%%%%%%%%%%%%%%%%%%%%%%%%%%%%%%%%%%%%%%%%%%%%%%%%%%%%%%%%%%%%%%%%%%
\begin{abstract}
Path planning algorithms, such as the search-based A*, are a critical component of autonomous mobile robotics, enabling robots to navigate from a starting point to a destination efficiently and safely. We investigated the resilience of the A* algorithm in the face of potential adversarial interventions known as obstacle attacks. The adversary's goal is to delay the robot's timely arrival at its destination by introducing obstacles along its original path.

We developed malicious software to execute the attacks and conducted experiments to assess their impact, both in simulation using TurtleBot in Gazebo and in real-world deployment with the Unitree Go1 robot.
% More than 90\% of the attacks successfully delayed or rerouted the robot in simulation, while all attacks deployed on real robots were successful. The percentage delay in all experiments was 32\% to 36\% on average. The results show that the algorithm's robustness is not solely an attribute of its design but is significantly influenced by the operational environment.

In simulation, the attacks resulted in an average delay of 36\%, with the most significant delays occurring in scenarios where the robot was forced to take substantially longer alternative paths. In real-world experiments, the delays were even more pronounced, with all attacks successfully rerouting the robot and causing measurable disruptions. These results highlight that the algorithm's robustness is not solely an attribute of its design but is significantly influenced by the operational environment. For example, in constrained environments like tunnels, the delays were maximized due to the limited availability of alternative routes.
 
\end{abstract}

%%%%%%%%%%%%%%%%%%%%%%%%%%%%%%%%%%%%%%%%%%%%%%%%%%%%%%%%%%%%%%%%%%%%%%%%%%%%%%%%
\section{INTRODUCTION}
With the recent focus on Artificial Intelligence (AI) across several fields, the adoption and development of robots and other AI-based cyber-physical systems have become much faster and easier, leading to widespread use of them.

While these technological advances have been significant, the focus has primarily been on functionality and innovations, leaving safety and security considerations, particularly in robotics, considerably behind. The environments where the robots are being deployed, such as logistics, public spaces and streets, and disaster areas, are rarely threat-free and can be subject to attacks from adversaries. While some studies have been conducted, the vulnerability of mobile robots to adversarial attacks remains largely underexplored. The need for cyber-security and cyber-safety in robotics differs from securing traditional computing systems due to the added dimension of physical interaction~\cite{Portugal2017}. This added aspect means that gaining control over a robot could result in harm, not only to the robot itself but also to its immediate environment, posing a potential and serious risk of injury to nearby individuals. 
%Furthermore, there is an absence of established norms and legal frameworks to address critical issues like robot accident accountability and appropriate autonomy levels near humans ~\cite{FoschVillaronga2019}. 

\begin{figure}[h]
    \centering
    \includegraphics[width=0.9\linewidth]{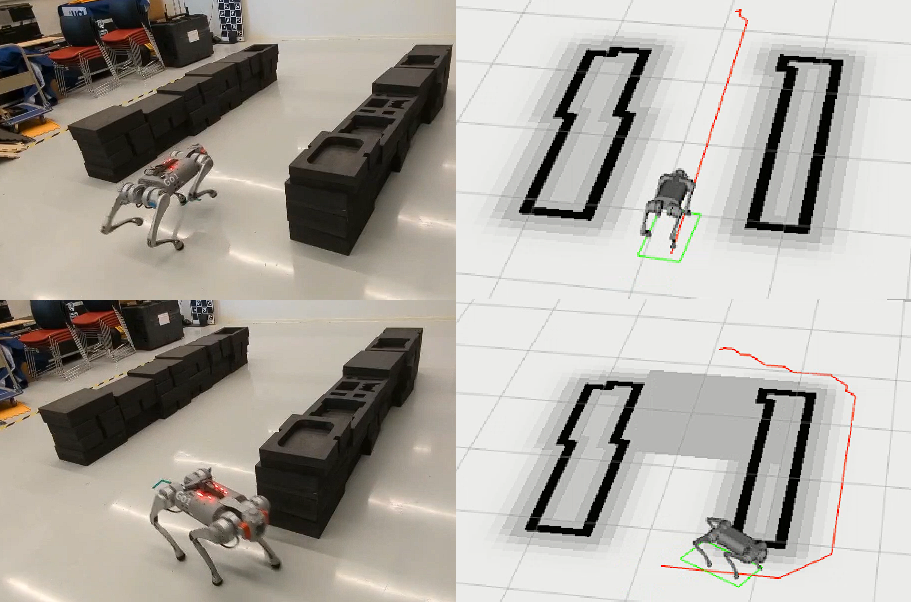}
    \caption{Real-world deployment without any attacks (top) and with the obstacle attack (bottom), showing the planned path in red. The attack does not change the physical environment, only the robot's cost map.}
    \label{fig:intro-figure}
\end{figure}

Given the threats faced by autonomous mobile robots in their operational environments, it is critical to develop robust path planning algorithms that can withstand adversarial manipulations. This study examines the robustness of search-based A* path planning against adversarial manipulations in the operational environment. It explores the implications for safety and security in autonomous robotics using a methodologically simple yet effective `obstacle attack'. While previous studies~\cite{vemprala2021adversarial,cavorsi2023multi,Wu2024} have focused on completely immobilising or severely restricting the agent's movement, we focus on causing delays in the agent's path-finding efforts, thereby reducing its likelihood of detection by defensive mechanisms. This is achieved by determining obstacles' optimal placement and subsequent integration into the operational environment to delay the robot's timely arrival at its destination. We conducted a set of experiments, both simulation within a static warehouse setting and on a physical robot in small environments, and collected data to demonstrate the effectiveness of the attacks. We studied the time required for the robotic agent to reach its goal and the strategic placement of obstacles.

The paper is organised as follows: Section~\ref{sec:literature-review} presents a comprehensive literature review, discussing a variety of adversarial attacks targeting robotic agents and the existing countermeasures; Section~\ref{sec:methodology} outlines the threat model and attack used in this study, describing the experimental setup for both simulation and real-world deployment, as well as the data collection techniques; Section~\ref{sec:results} details the data gathered from the experiments, providing a quantitative foundation for subsequent analysis; Section~\ref{sec:discussion} provides an interpretation of the data with conclusions regarding the algorithm's resilience; and Section~\ref{sec:conclusion} concludes the paper, summarising the research findings, drawing implications for future developments in autonomous robotics.
%, and proposing directions for future research.

\section{RELATED WORK}\label{sec:literature-review}
Path planning has been heavily studied in the past~\cite{ellis2022navigation, liu2023vit, liu2024dipper, stamatopoulou2024dippest}. Here, we briefly review the work related to adversarial attacks in robots. 

While there have been some studies on the security threats that affect robotic hardware, firmware/OS, and applications~\cite{Clark2017,Creado2017,AhmadYousef2018,Yaacoub2021}, less focus has been given to studying the vulnerabilities of AI deployed in robotics~\cite{Calo2018}. Robots interacting with a physical environment bring a new category of threats beyond software and hardware risks referred to as behaviour threats~\cite{Colledanchise2021}. For example, an autonomous delivery robot can make nearby pedestrians feel threatened, prompting them to take action to render the robot immobile~\cite{Oravec2022}. %These risks encompass various forms, such as exploiting a robot's authorised access, manipulating its sensor systems, and intentionally inducing harmful actions in the robot.

Adversarial attacks can significantly impact autonomous robots, disrupting control systems and decision-making capabilities. Vemprala et al.~\cite{vemprala2021adversarial} investigate the susceptibility of iterative optimisation-based trajectory planners to adversarial attacks, showing that such attacks can disrupt the optimisation process by manipulating the cost function and lead to failures or increased computation times. The authors test their method by applying adversarial strategies to two state-of-the-art trajectory planners, demonstrating that these planners are consistently vulnerable to such attacks. Cavorsi et al.~\cite{cavorsi2023multi} present an algorithm for resilient path planning in robot teams, where robots can coordinate despite adversaries. The authors use Control Barrier Functions (CBFs) to balance resilience and safe navigation, developing a method to identify and handle trade-offs between these objectives. Wu et al.~\cite{Wu2024} focus on the vulnerability of robot motion planning algorithms to adversarial attacks, specifically in the physical environment. Their study highlights two types of attacks: planner failure and blindspot attacks. Planner failure attacks involve making subtle changes to the physical environment, causing the motion planner to fail to find a valid path. In contrast, blindspot attacks exploit occlusions and sensor field-of-view limitations, leading the planner to generate a collision-free trajectory that intersects with unperceived obstacles. The reported experimental results indicate that these attacks can drastically increase failure rates to $95\%$ and collision rates to $90\%$ with only minor environmental modifications. 
% The paper also analyses the transferability of these attacks across different planning algorithms, underscoring the serious threat they pose to the reliability of robotic systems. Jia et al.~\cite{Jia2022} focus on object location identification, using adversarial patches to mislead detectors. Their experiment with an industrial robotic arm demonstrates the risk of misidentifying objects, illustrating the tangible hazards posed by adversarial attacks.

Clark et al.~\cite{Clark2018} examine the manipulation of an autonomous robotic vehicle's ultrasonic collision avoidance sensors to redirect its path. By employing the Q-learning algorithm for real-time route control, they illustrate the potential for indirect attacks on robotic systems, emphasising the need for robust countermeasures. 
% Sharma et al.~\cite{Sharma2019} study adversarial attacks on connected autonomous vehicles using the VeReMi dataset. Their approach involves training adversarial networks to produce realistic examples, revealing vulnerabilities in networked vehicular systems and highlighting the complexity of ensuring their security. Melis et al.~\cite{Melis2017} investigate the susceptibility of iCub robot-vision systems to adversarial images using support vector machines (SVMs) on the iCubWorld dataset. They propose a countermeasure involving rejecting abnormal inputs and enhancing system robustness against attacks. Otomo et al.~\cite{Otomo2022} target legged robots like the quadruped Ant-v2 and bipedal Humanoid-v2, applying adversarial torque perturbations to induce falls. Their study reveals vulnerabilities in the quadruped robot, while the bipedal robot demonstrates robustness, offering insights into diagnosing walking instability in robotics.
Shi et al.~\cite{Shi2024} explore the vulnerabilities of learning-based locomotion controllers in quadrupedal robots using adversarial attacks. Their study focuses on the robustness of controllers that employ deep Reinforcement Learning (RL) against real-world uncertainties like sensor noise and external perturbations. Despite empirical robustness, this research shows that these controllers can fail significantly under carefully crafted, low-magnitude adversarial sequences. 
% Through experiments conducted in both simulation and real-world settings, Shi et al. validate their approach, demonstrating its effectiveness in identifying weaknesses in locomotion controllers. The insights gained from these attacks can be used to enhance the original policies and improve the safety of these systems, underscoring the importance of thorough testing and validation in developing robust robotic controllers.

% Defensive strategies and robustness evaluation are crucial for protecting machine learning models from adversarial attacks, especially in robotics and autonomous systems. 

Lechner et al.~\cite{Lechner2021} studied the robustness-accuracy trade-off, exploring the balance between maintaining accuracy and ensuring robustness against adversarial attacks. Focusing on two neural controllers - one for vision-based commands and another for LiDAR-based motor control; their findings suggest that adversarial training can introduce new errors, potentially undermining the robot's learned behaviour. Lechner et al.~\cite{Lechner2023} revisit the robustness-accuracy trade-off, incorporating advancements like overparametrisation and vision transformers. Their study shows that integrating adversarial weight perturbation reduces the gap between robustness and accuracy.

% Melis et al.~\cite{Melis2017} propose an efficient countermeasure for iCub robot-vision systems, rejecting abnormal inputs within "blind spots." This enhances system defences against adversarial attacks, demonstrating a practical resilience-building approach. Zizzo et al.~\cite{Zizzo2019} explore adversarial attacks on intrusion detection systems (IDS) within industrial control systems (ICS). They recommend adversarial training and additional detection systems as defences, highlighting parallels between ICS and robotics in handling sensor data.

% Exploring adversarial machine learning extends beyond immediate threats to consider broader implications and future research directions.
% Kuutti et al.~\cite{Kuutti2020} introduce automated black-box testing for deep control policies using adversarial reinforcement learning. The adversarial agent, trained with reinforcement learning, is placed in the testing environment to degrade the performance of the target agent and cause it to collide. Testing their approach on an autonomous vehicle, they discovered vulnerabilities that were missed during online testing and highlighted the benefit of automated testing over manual methods.

% They propose potential future research avenues, such as data augmentation and novel training methods, to bridge this gap fully.

To the best of our knowledge, no previous studies have examined the robustness of path planning algorithms under adversarial manipulations in operational environments where obstacles are introduced along the intended path to delay timely arrival at the destination.

\section{METHODOLOGY} 
\label{sec:methodology}
In this section, the threat model will first be introduced, covering the adversary's objectives, incentives, knowledge, and capabilities. It then introduces the obstacle attack and experimental setup to show how the attacks were implemented in simulation and validated real-world deployments, along with the data collection process. 

\begin{figure*}[h]
    \centering
    \includegraphics[width=0.9\linewidth]{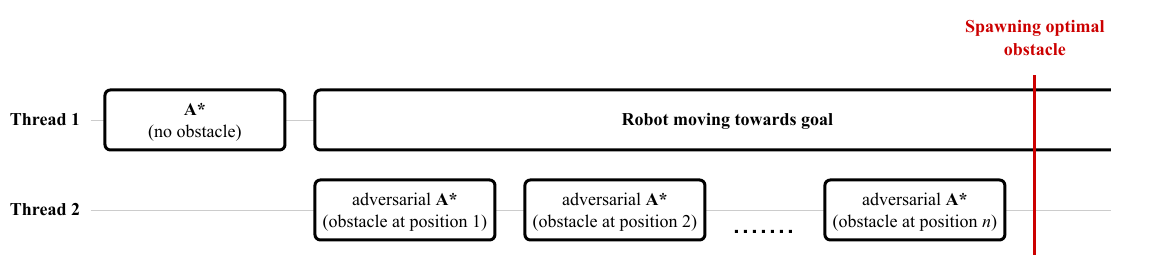}
    \caption{Parallel execution of the obstacle attack.}
    \label{fig:attack-thread}
\end{figure*}

\subsection{Threat Model}
\textit{Objectives:} The adversary's objective is to delay the physical agent's timely arrival at its destination by introducing obstacles along its original path, whether physical or virtual, thereby forcing it to take less optimal alternative routes. If necessary, the adversary could modify the attack to direct the agent along a specific path or through a particular point on the map, transforming the initial non-targeted attack into a targeted one. The adversary's goal is to ensure that the attack remains as undetectable as possible, limiting it to adding a single obstacle.

\textit{Incentives:} The adversary's motivations predominantly hinge on inflicting financial harm. This includes tactics such as delaying or diverting delivery robots for potential theft opportunities and inflicting reputational damage on companies that operate robotic systems or directly on the robotic systems themselves. Furthermore, these malicious actions could extend to risking human lives, especially concerning robots employed in critical roles like disaster relief and rescue services. The extent of these actions not only threatens the operational integrity and trust in robotic solutions but also emphasises the need for robust security measures and ethical guidelines to mitigate such risks and safeguard both property and lives.

\textit{Knowledge \& Capability:} We assume that the adversary could carry out the attack virtually by manipulating the agent's cost map to create a false obstacle. This manipulation could be achieved through a man-in-the-middle attack~\cite{conti2016survey} or malicious software. To calculate the optimal obstacle placement, the adversary would need access to the map of the environment in which the agent operates. We assume that the adversary possesses this knowledge, as well as the agent's starting (current) and goal positions. These assumptions are reasonable, given that robots repeatedly operate in the same environment.

\subsection{Obstacle Attack}
The attack described employs a method we named brute-force obstacle attack. This approach methodically tests every feasible location for placing an obstacle along the path initially planned. The process is summarised by pseudo-code in Algorithm~\ref{alg:attack}.

\begin{algorithm}[ht]
\caption{Obstacle Attack}
\label{alg:attack}
\begin{algorithmic}[1]
    \STATE $path \gets \textsc{A*}(start, goal, map)$
    \STATE $path_{longest} \gets path$
    \FOR{$step$ \textbf{in} $path$}
        \STATE $path_{new} \gets$ run A* with obstacle at $step$
        \IF{$\textsc{Length}(path_{new}) > \textsc{Length}(path_{longest})$}
            \STATE $obstacle_{best} \gets$ obstacle at $step$
            \STATE $path_{longest} \gets path_{new}$
        \ENDIF
    \ENDFOR
    \STATE add $obstacle_{best}$ to $map$
\end{algorithmic}
\end{algorithm}

The process begins by determining the actual path that the agent should follow. This is achieved by deploying the global path planning algorithm, in our case A*, which requires the starting point, destination (goal), and cost map as inputs. Following this initial use of the A* algorithm to establish the baseline path, the algorithm is executed repeatedly, equivalent to the number of steps identified within the original path. During these subsequent iterations, an obstacle is introduced into the cost map at different positions. This technique makes it feasible to ascertain the most strategic placement of the obstacle that yields the maximum path extension required for the agent to reach its goal.

The size of the obstacle is carefully chosen to balance two objectives: the obstacle must be large enough to force the agent to take a significantly longer alternative route, but it should not be so large as to completely block the agent's path, which could render the scenario unrealistic or trivial.

The location of the obstacle is determined through a brute-force search across all possible positions along the agent's path. For each position, the algorithm evaluates the resulting path extension and selects the placement that maximizes the delay. This iterative process ensures that the obstacle is positioned in the most disruptive yet plausible location, aligning with the adversary's goal of maximizing impact while maintaining undetectability.
   
Once the optimal position for the obstacle has been identified through this iterative process, the obstacle is inserted in the agent's cost map, thus impacting the path's overall length and complexity. The effectiveness of this method lies in its simplicity and the brute-force nature of testing every possible variation to find the most obstructive placement of the obstacle.

For the A* algorithm, the larger the map, the longer it takes to find a feasible path. Running the obstacle attack algorithm sequentially would, therefore, cause a delay between receiving the goal position and commencing the motion towards it. To address this, the attack algorithm has been designed to run in parallel with the normal function of the Robot Operating System (ROS) \cite{288} node. Fig.~\ref{fig:attack-thread} illustrates the two threads utilised in the attack process. Thread 1 encompasses the intended sequence of execution, wherein it first determines the path to the goal and then initiates the robot's movement along the specified path towards the goal. Thread 2 is commenced after the path planning algorithm is completed. It is where the adversarial A* algorithm runs once for each position along the path, and upon evaluating the last position, it deploys the obstacle in the optimal position. As thread 2 runs concurrently with the robot's movement, the robot may have already traversed the obstacle position; this is one type of data we evaluate in the experiments.

% \begin{figure*}[h]
%     \centering
%     \includegraphics[width=\linewidth]{Figures/obstacle_attack_threads.drawio.pdf}
%     \caption{Parallel execution of the obstacle attack.}
%     \label{fig:attack-thread}
% \end{figure*}

\subsection{Experimental Setup}

In the following sections, we explain how the attacks were created first through simulation using path planning simulated in Gazebo and later validated using real-world deployment on a Unitree Go1 robot~\footnote{https://www.unitree.com/go1/}. 

\subsubsection{Simulation}
The experiments were conducted in Gazebo simulation environment using the ROS. An existing project has been selected as a reference for path planning simulated in Gazebo. The project features a ROS implementation of various path planning algorithms, a simulated warehouse environment, and a robotic agent. Specifically, we used the TurtleBot 3 Waffle as our robotic agent, equipped with the 360 Laser Distance Sensor LDS-01, a 2D laser scanner capable of 360-degree sensing. This sensor collects data around the robot for use in SLAM. Our global path planning algorithm was A*, while we employed a PID controller for local path planning. To carry out the attack, we deployed a malicious ROS node as the attack vector.

\begin{figure}[h]
    \centering
    \includegraphics[width=\linewidth,trim={8cm 2cm 9cm 3cm},clip]{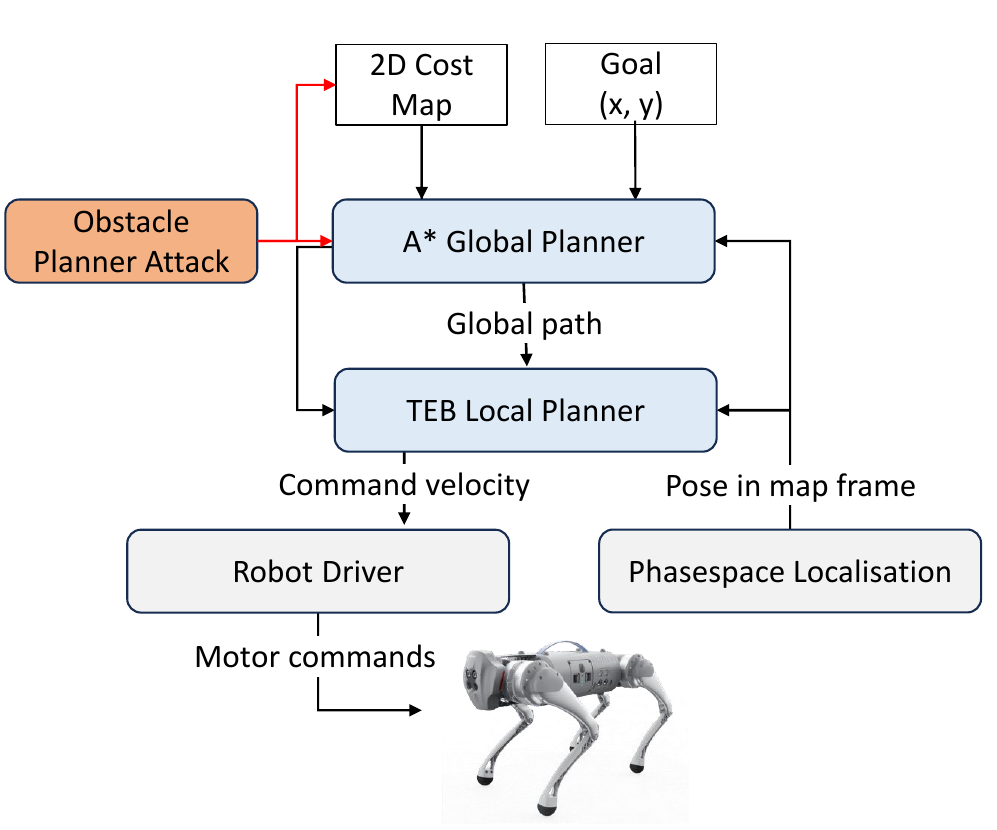}
    \caption{Navigation stack for the real robot deployment.}
    \label{fig:navigation-stack}
\end{figure}

\subsubsection{Real-world Deployment}
A schematic representation of the real-world deployment of the obstacle attack is depicted in Fig.~\ref{fig:navigation-stack}. We validate the performance of the attack in real-world conditions on a Unitree Go1. To leverage the existing robot navigation frameworks, the attack is integrated with the 2D ROS navigation stack to act as a part of the global path planner. Given the cost map, the attack generates an obstacle in the global path, making sure the obstacle does not fully block all the possible paths. The local planner and obstacle avoidance algorithm used is the Timed-Elastic-Band (TEB)~\cite{rosmann2012trajectory, rosmann2013efficient} with localisation provided by the Phasespace tracking system in real-time at $960$ Hz. Examples of successful deployment of the pipeline can be seen in Fig.~\ref{fig:intro-figure}.

\subsection{Data Collection}
To collect data, we conducted the experiments in simulation and real-world deployment. \textbf{Simulation scenario}: We positioned the robot in the centre of the warehouse and instructed it to travel to one of the $23$ specified locations, where the focus was on end-to-end navigation in larger environments. 
\textbf{Real-world scenarios}: We conducted three scenarios with a real robot, each scenario with a different map (tunnel, turn, and t-wall), testing the robot's performance in simpler, smaller environments. 

Both the simulation and real-world experiments were repeated three times for each map/start/end without any interference (known as benign conditions) and three times for each map/start/end with interference (known as adversarial conditions), resulting in $157$ experimental runs. The data we collected included the coordinates of the start and end locations, the time taken by the robot to reach them, and for runs with interference, the time taken to introduce the obstacle, the coordinates of the obstacle, and whether the obstacle was placed before the robot reached the obstacle's location.

\section{RESULTS}\label{sec:results}
This section presents the results from the data collection for both simulated and real-world experiments.

\begin{figure}[h]
    \centering
    \includegraphics[width=\linewidth]{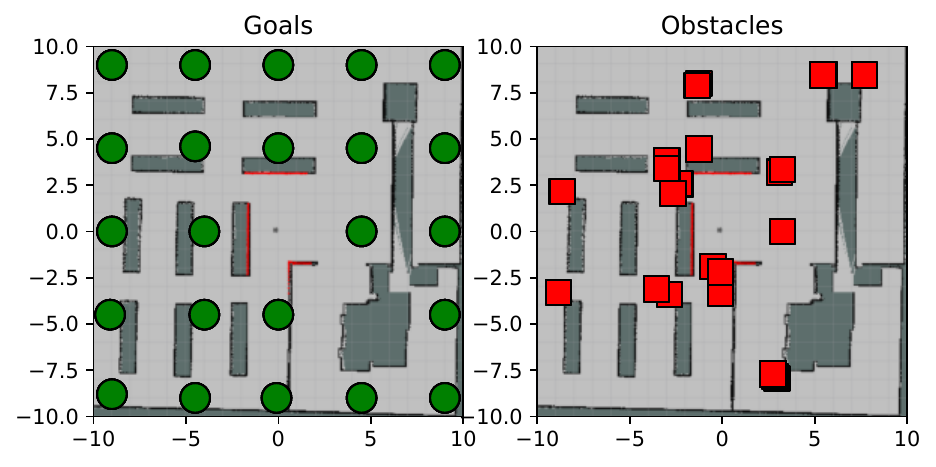}
    \caption{Goal positions used in the experiments (left) and all the calculated obstacle positions (right).}
    \label{fig:goals_and_obstacles}
\end{figure}

Fig.~\ref{fig:goals_and_obstacles} compares simulation scenarios with predetermined goals and dynamically placed obstacles, where it can be seen that the attack places obstacles in narrow pathways and on the corners to delay the robot the most.

\begin{figure}[h]
    \centering
    \includegraphics[width=0.8\linewidth]{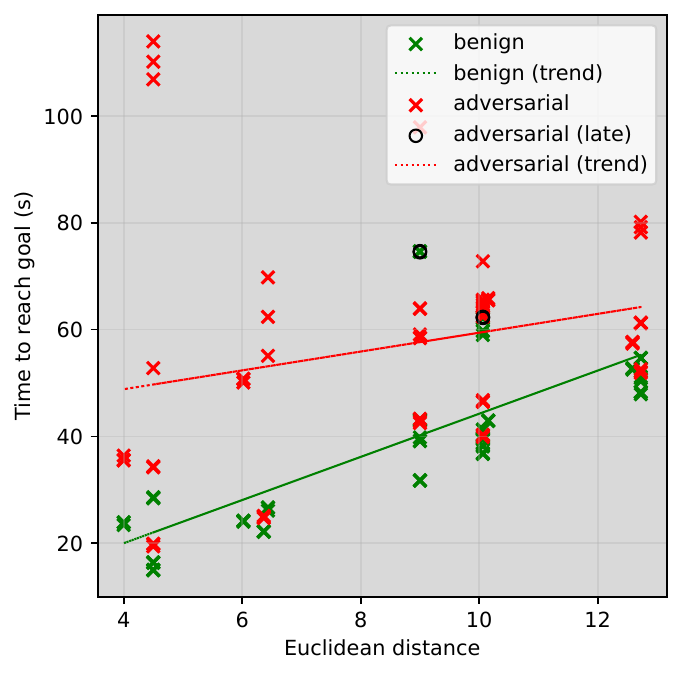}
    \caption{A comparison of robotic navigation times under benign and adversarial conditions, showing the impact of obstacles on travel time relative to Euclidean distance.}
    \label{fig:scatter-data}
\end{figure}

Fig.~\ref{fig:scatter-data} presents the simulation results comparing the performance of a robotic system under benign and adversarial conditions. The central scatter plot illustrates the relationship between the Euclidean distance to a goal and the time taken to reach that goal, capturing all the data points from the experiment. In the benign condition, the robot navigates through the environment with no interference from the adversary. The data points exhibit a clear upward trend, where the time to reach the goal increases with the Euclidean distance. In contrast, in the adversarial condition, the robot encounters an obstacle that complicates its path, affecting its efficiency. The adversarial trend is less steep than the benign trend, suggesting that although obstacles introduce delays, the robot can somewhat mitigate the impact on travel time across varying distances, opening the possibility of adaptive path finding strategies that limit the increase in time as distance grows.

A critical factor influencing the success of the attack is the timing of obstacle injection relative to the agent's progress along its path. Obstacles injected early in the path give the agent more opportunities to replan efficiently, often resulting in smaller delays. For example, if an obstacle is placed near the start of the path, the agent can explore alternative routes with minimal backtracking, reducing the overall impact of the attack.

% Conversely, obstacles injected late in the path, particularly immediately before the agent reaches the targeted location, tend to cause more significant delays. In these cases, the agent must backtrack and replan from a point closer to the destination, leading to longer detours and increased travel time. This scenario represents the worst-case performance for the agent, as it maximizes the disruption caused by the adversarial obstacle.

\begin{figure}[h]
    \centering
    \includegraphics[width=0.9\linewidth]{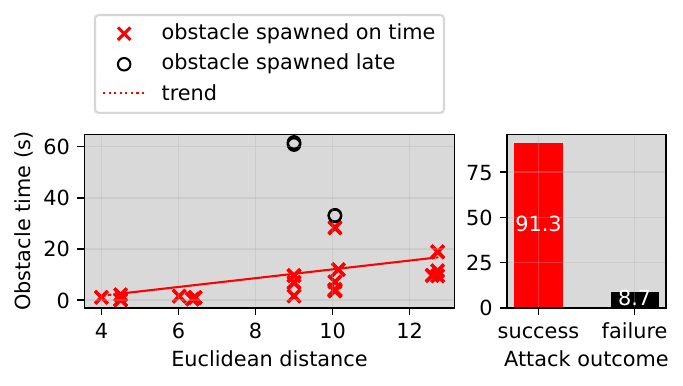}
    \caption{The relationship between Euclidean distance, obstacle spawn time, and the success rate of obstacle attacks.}
    \label{fig:scatter-spawn-time}
\end{figure}

Fig.~\ref{fig:scatter-spawn-time} consists of two plots that analyse the timing and success of obstacle spawning in the simulated experiments. The high success rate of $91.3\%$ indicates that the adversarial A* algorithm is generally effective at identifying optimal injection times to maximize delays. However, the $8.7\%$ failure rate highlights instances where obstacles were injected too late or in suboptimal locations, allowing the agent to bypass them with minimal disruption. These failures underscore the importance of precise timing in adversarial attacks and suggest that further refinement of the injection strategy could improve the algorithm's effectiveness.

\begin{figure}[h]
    \centering
    \includegraphics[width=0.78\linewidth]{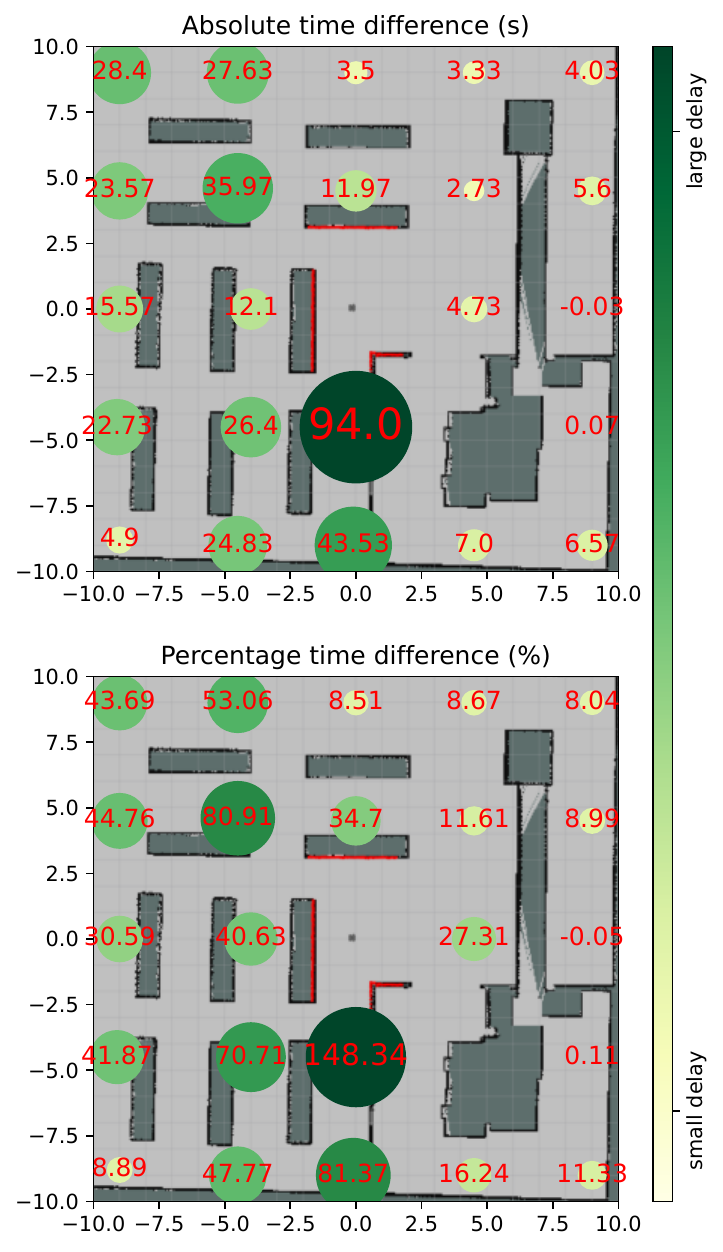}
    \caption{Comparison of mean absolute and percentage difference in time taken between benign and adversarial scenarios for each goal position.}
    \label{fig:time_delays}
\end{figure}

Fig.~\ref{fig:time_delays} presents a detailed comparison of the performance of the A* path planning algorithm under two distinct conditions: ``benign'' navigation and ``adversarial'' navigation, where obstacles were intentionally introduced as part of an obstacle attack strategy. The figure captures both the mean absolute (top subfigure) and percentage differences (bottom subfigure) in navigation time between these two scenarios, providing insights into the effectiveness of the obstacle attacks. The mean time delay across all goal positions is $17.78$ seconds, while the percentage delay across all experiments is $36.01\%$.

\begin{figure}[h]
    \centering
    \begin{subfigure}[b]{0.4\linewidth}
        \centering
        \includegraphics[width=\linewidth,trim={0cm 0.3cm 0cm 0.4cm},clip]{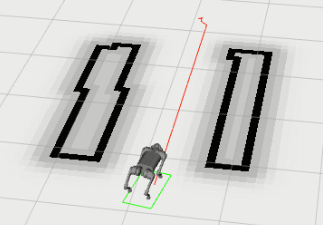}
        \caption{}
        \label{fig:tunnel-0}
    \end{subfigure}
    \begin{subfigure}[b]{0.4\linewidth}
        \centering
        \includegraphics[width=\linewidth]{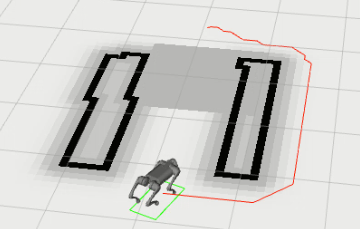}
        \caption{}
        \label{fig:tunnel-1}
    \end{subfigure}
    \begin{subfigure}[b]{0.4\linewidth}
        \centering
        \includegraphics[width=\linewidth]{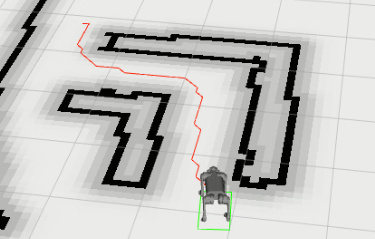}
        \caption{}
        \label{fig:turn-0}
    \end{subfigure}
    \begin{subfigure}[b]{0.4\linewidth}
        \centering
        \includegraphics[width=\linewidth,trim={0cm 0.4cm 0cm 1cm},clip]{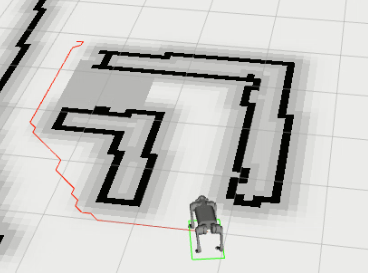}
        \caption{}
        \label{fig:turn-1}
    \end{subfigure}
    \begin{subfigure}[b]{0.4\linewidth}
        \centering
        \includegraphics[width=\linewidth,trim={3cm 1cm 3cm 1cm},clip]{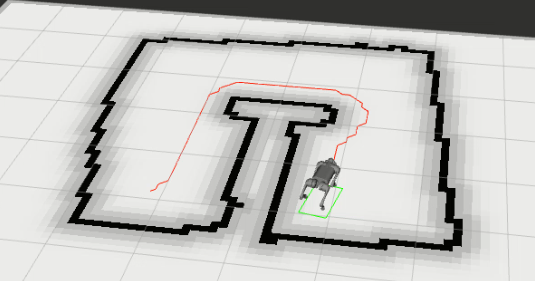}
        \caption{}
        \label{fig:t-wall-0}
    \end{subfigure}
    \begin{subfigure}[b]{0.4\linewidth}
        \centering
        \includegraphics[width=\linewidth,trim={3cm 1.35cm 3cm 1cm},clip]{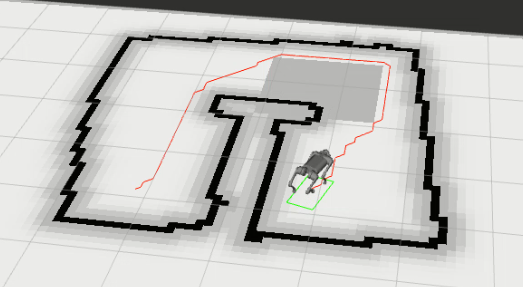}
        \caption{}
        \label{fig:t-wall-1}
    \end{subfigure}
    \caption{The 3 maps used in real-world experiments: tunnel (a)(b), turn (c)(d), and t-wall (e)(f). The first column shows the original path taken without an attack, while the second column shows the calculated obstacle placement (grey square) and the alternative path taken by the robot.}
    \label{fig:real-world_experiemtns}
\end{figure}

The real-world experiments show performance on smaller-scale maps compared to the simulation. This was to investigate the effects of the obstacle attack on robots operating in environments with limited mobility. The maps used in these experiments can be seen in Fig.~\ref{fig:real-world_experiemtns}, and it shows both the benign and adversarial scenarios.

% The results are summarised in Table~\ref{tab:experimental_results}. The attack caused the biggest delay when the robot was ordered to go straight through a tunnel; the obstacle completely redirected the robot and caused a $10$-second delay on average. The robot was also redirected to the second environment, where it had to turn, but the delays were minimal, with the time taken being almost the same as without the attack. In the last case, the robot was not fully redirected due to there not being a different path, but it still managed to delay the robot's arrival time by almost $10\%$.

The results are summarized in Table~\ref{tab:experimental_results}. The attack caused the largest delay in the tunnel scenario, where the obstacle completely redirected the robot, resulting in an average delay of $10$ seconds. This scenario represents the worst-case performance for the agent, as the constrained environment limited the robot's ability to find efficient alternative paths. In the turn scenario, the delays were minimal, with the robot's travel time being almost the same as without the attack. This suggests that the impact of adversarial obstacles depends heavily on the environment's layout and the availability of alternative routes. In the final scenario, where the robot was not fully redirected due to the lack of alternative paths, the attack still managed to delay the robot's arrival time by almost $10\%$. This demonstrates that even in less ideal conditions for the adversary, the A* algorithm remains susceptible to delays caused by strategically placed obstacles.

% \begin{table}[h]
%     \centering
%     \begin{tabular}{|c|c|c|c|c|} \hline
%          \textbf{Scenario} & \textbf{Map} & \makecell{\textbf{Success}\\\textbf{rate}} & \makecell{\textbf{Mean}\\\textbf{absolute}\\\textbf{delay}} & \makecell{\textbf{Mean}\\\textbf{percentage}\\\textbf{delay}} \\ \hline %\hline
%          simulation & warehouse & 91.3\% & \textbf{17.78s} & 36.01\% \\ \hline
%          \multirow{4}{4em}{real-world} & tunnel & \textbf{100\%} & 10.07s & \textbf{86.08\%} \\ \cline{2-5}
%          & turn & \textbf{100\%} & 0.23s & 1.75\% \\ \cline{2-5}
%          & t-wall & \textbf{100\%} & 1.57s & 9.65\% \\ \cline{2-5}
%          & \textit{overall} & \textbf{\textit{100\%}} & \textit{3.95s} & \textit{32.49\%} \\ \hline
%     \end{tabular}
%     \caption{Experimental results.}
%     \label{tab:experimental_results}
% \end{table}

\begin{table}[h]
    \centering
    \begin{tabular}{|c|c|c|c|c|} \hline
         \textbf{Scenario} & \textbf{Map} & \makecell{\textbf{Mean}\\\textbf{absolute}\\\textbf{delay}} & \makecell{\textbf{Mean}\\\textbf{percentage}\\\textbf{delay}} & \makecell{\textbf{Success}\\\textbf{rate}} \\ \hline %\hline
         simulation & warehouse & \textbf{17.78s} & 36.01\% & 91.3\% \\ \hline
         \multirow{4}{4em}{real-world} & tunnel & 10.07s & \textbf{86.08\%} & \textbf{100\%} \\ \cline{2-5}
         & turn & 0.23s & 1.75\% & \textbf{100\%} \\ \cline{2-5}
         & t-wall & 1.57s & 9.65\% & \textbf{100\%} \\ \cline{2-5}
         & \textit{overall} & \textit{3.95s} & \textit{32.49\%} & \textbf{\textit{100\%}} \\ \hline
    \end{tabular}
    \caption{Experimental results.}
    \label{tab:experimental_results}
\end{table}

\section{DISCUSSION}\label{sec:discussion}
% The high success rate ($91.3\%$ in simulation and $100\%$ in real-world scenarios) of the obstacle attacks succeeded in delaying the robot's progress. This effectively shows the vulnerabilities of the current search-based A* path planning algorithm to adversarial conditions. Such a high success rate underscores the urgent need for more robust path planning algorithms capable of withstanding these attacks. This observation sheds light on the areas of vulnerability within robotic path planning and navigation, prompting a reevaluation of existing strategies and encouraging the pursuit of more resilient solutions.

The experiments revealed that adversarial obstacles can cause significant delays, with the worst-case delays reaching up to $10$ seconds in the tunnel scenario. These delays highlight the vulnerabilities of the current search-based A* path planning algorithm to adversarial conditions. While the success rate of the attacks was high ($91.3\%$ in simulation and $100\%$ in real-world scenarios), the magnitude of the delays provides a more meaningful measure of the algorithm's susceptibility. For example, in environments with limited alternative routes, such as tunnels, the delays were maximized, underscoring the algorithm's dependence on environmental factors. These findings emphasize the urgent need for more robust path planning algorithms capable of withstanding adversarial attacks.

From the trends associated with mean absolute and percentage delays, we observed that traversal times increase with distance under benign and adversarial conditions. However, it is noteworthy that the trend line is less steep in adversarial conditions, signalling a relatively uniform impact of obstacles irrespective of distance. This observation led us to speculate on the nature of the challenges faced by the robot in navigational tasks. Despite the introduction of obstacles presenting a consistent difficulty level, our findings suggest that the robot's performance does not significantly deteriorate over longer distances when obstacles are present. This could imply the existence of a "threshold effect", where beyond a certain level of introduced complexity, further increments in distance do not correspond with a disproportionate increase in navigational difficulty. 

% Significant variability was apparent in our observation of navigation performance across a range of goal positions, as depicted by the differing sizes and colours of circles within our visualisation. This variability underscores a critical point for discussion: the pressing need to understand how distinct environmental factors, alongside the positioning of goals, significantly influence navigation outcomes. Such insights highlight the potential benefits of adopting personalised or context-aware path planning strategies. These tailored approaches could enhance navigation efficiency across many scenarios, suggesting a paradigm shift from a one-size-fits-all methodology to more nuanced and adaptable navigation solutions. This tailoring to individual or scenario-specific requirements could be the key to optimising navigational tasks in environments with varied challenges to navigational systems.

Another important observation was the high impact of the attack strategies in constrained spaces, specifically on the left side of the simulated map. This area, having limited space for motion, saw significant delays. The restricted manoeuvrability in such areas severely compromises the robot's ability to adapt to and circumvent obstacles, leading to notable increases in delay times. By carefully considering the spatial characteristics of the environment, one can maximise the impact of adversarial conditions, pointing to a nuanced understanding of how space and adversarial strategies interact. 

\section{CONCLUSION}\label{sec:conclusion}
We investigated the resilience of the search-based path planning algorithm A* in the face of potential adversarial interventions known as obstacle attacks. We introduced the threat model, defined obstacle attack as a brute-force effort and tested its efficacy through both simulation and real-world deployment. The findings shows that the algorithm's robustness is not solely an attribute of its design but is significantly influenced by the operational environment. Specifically, factors such as the size of the environment and the proximity to the designated goal played crucial roles. 
% Interestingly, the obstacle attack demonstrated a high success rate, impacting the robotic agent in over 90\% of the test cases. However, it's important to note that a considerable percentage of these instances caused minimal delays.

While this study focuses on adding a single obstacle to simplify the analysis, the approach is justified by the need to establish a baseline understanding of the adversarial impact. A single obstacle provides a clear and measurable way to evaluate the worst-case delay without introducing additional variables, such as varying obstacle sizes or quantities. However, we acknowledge that in real-world scenarios, an adversary might employ multiple smaller obstacles to achieve a similar delay while maintaining a lower profile. For example, several small obstacles could be strategically placed to create a cumulative delay without being as visually conspicuous as a single large obstacle.

% This trade-off between obstacle size, quantity, and detectability warrants further exploration in future work. For example, a comparative analysis of single versus multiple obstacles could reveal how different attack strategies balance effectiveness and stealth. Such an investigation would provide deeper insights into the practical implications of adversarial attacks in real-world systems, where detectability is a critical concern.

This research contributes to a better understanding of the vulnerabilities inherent in the widely used A* path planning algorithm. It demonstrates the importance of considering environmental variables in assessing algorithmic robustness. Furthermore, the successful execution of the obstacle attack in many test scenarios highlights a critical area for future research: improving the security of robotic systems by developing more resilient algorithms to protect autonomous systems against such adversarial manipulations. 

%                                   % the textheight of the last page by a suitable amount.
%                                   % This command does not take effect until the next page
%                                   % so it should come on the page before the last. Make
%                                   % sure that you do not shorten the textheight too much.

%%%%%%%%%%%%%%%%%%%%%%%%%%%%%%%%%%%%%%%%%%%%%%%%%%%%%%%%%%%%%%%%%%%%%%%%%%%%%%%%

%%%%%%%%%%%%%%%%%%%%%%%%%%%%%%%%%%%%%%%%%%%%%%%%%%%%%%%%%%%%%%%%%%%%%%%%%%%%%%%%



% \section*{ACKNOWLEDGMENT}

% The preferred spelling of the word ÒacknowledgmentÓ in America is without an ÒeÓ after the ÒgÓ. Avoid the stilted expression, ÒOne of us (R. B. G.) thanks . . .Ó  Instead, try ÒR. B. G. thanksÓ. Put sponsor acknowledgments in the unnumbered footnote on the first page.

%%%%%%%%%%%%%%%%%%%%%%%%%%%%%%%%%%%%%%%%%%%%%%%%%%%%%%%%%%%%%%%%%%%%%%%%%%%%%%%%
\bibliographystyle{IEEEtran}
\bibliography{IEEEabrv, references}

@inproceedings{rosmann2012trajectory,
  title         = {{Trajectory Modification Considering Dynamic Constraints of Autonomous Robots}},
  author        = {R{\"o}smann, Christoph and Feiten, Wendelin and W{\"o}sch, Thomas and Hoffmann, Frank and Bertram, Torsten},
  booktitle     = {German Conference on Robotics (ROBOTIK)},
  pages         = {1--6},
  year          = {2012},
  organization  = {VDE}
}

@inproceedings{rosmann2013efficient,
  title         = {{Efficient Trajectory Optimization Using a Sparse Model}},
  author        = {{C. R{\"o}smann, W. Feiten, T. W{\"o}sch, F. Hoffmann, and T. Bertram}},
  booktitle     = {European Conference on Mobile Robots},
  pages         = {138--143},
  year          = {2013}
}

@inproceedings{Portugal2017,
  author        = {D. Portugal and S. Pereira and M. S. Couceiro},
  title         = {The role of security in human-robot shared environments: A case study in ROS-based surveillance robots},
  booktitle     = {2017 26th IEEE International Symposium on Robot and Human Interactive Communication (RO-MAN)},
  organization  = {IEEE},
  year          = {2017},
  month         = {aug},
  doi           = {10.1109/roman.2017.8172422}
}

@inproceedings{Clark2017,
  author        = {G. W. Clark and M. V. Doran and T. R. Andel},
  title         = {Cybersecurity issues in robotics},
  booktitle     = {2017 IEEE Conference on Cognitive and Computational Aspects of Situation Management (CogSIMA)},
  organization  = {IEEE},
  year          = {2017},
  month         = {mar},
  doi           = {10.1109/cogsima.2017.7929597}
}

@incollection{Creado2017,
  author        = {O. M. Creado and P. D. Le},
  title         = {Enforcing Security in Artificially Intelligent Robots Using Monitors (Short Paper)},
  booktitle     = {Lecture Notes in Computer Science},
  publisher     = {Springer International Publishing},
  pages         = {648--659},
  year          = {2017},
  doi           = {10.1007/978-3-319-72359-4_40}
}

@inproceedings{Clark2018,
  author        = {G. Clark and M. Doran and W. Glisson},
  title         = {A Malicious Attack on the Machine Learning Policy of a Robotic System},
  booktitle     = {2018 17th IEEE International Conference On Trust, Security And Privacy In Computing And Communications/ 12th IEEE International Conference On Big Data Science And Engineering (TrustCom/BigDataSE)},
  organization  = {IEEE},
  year          = {2018},
  month         = {aug},
  doi           = {10.1109/trustcom/bigdatase.2018.00079}
}

@article{AhmadYousef2018,
  author        = {K. Ahmad Yousef and A. AlMajali and S. Ghalyon and W. Dweik and B. Mohd},
  title         = {Analyzing Cyber-Physical Threats on Robotic Platforms},
  journal       = {Sensors},
  volume        = {18},
  number        = {5},
  pages         = {1643},
  year          = {2018},
  month         = {may},
  publisher     = {MDPI AG},
  doi           = {10.3390/s18051643}
}

@article{Calo2018,
  author        = {R. Calo and I. Evtimov and E. Fernandes and T. Kohno and D. O'Hair},
  title         = {Is Tricking a Robot Hacking?},
  journal       = {SSRN Electronic Journal},
  publisher     = {Elsevier BV},
  year          = {2018},
  doi           = {10.2139/ssrn.3150530}
}

@inproceedings{Lechner2021,
  author        = {M. Lechner and R. Hasani and R. Grosu and D. Rus and T. A. Henzinger},
  title         = {Adversarial Training is Not Ready for Robot Learning},
  booktitle     = {2021 IEEE International Conference on Robotics and Automation (ICRA)},
  organization  = {IEEE},
  year          = {2021},
  month         = {may},
  doi           = {10.1109/icra48506.2021.9561036}
}

@article{Yaacoub2021,
  author        = {J.-P. A. Yaacoub and H. N. Noura and O. Salman and A. Chehab},
  title         = {Robotics cyber security: vulnerabilities, attacks, countermeasures, and recommendations},
  journal       = {International Journal of Information Security},
  volume        = {21},
  number        = {1},
  pages         = {115--158},
  year          = {2021},
  month         = {mar},
  publisher     = {Springer Science and Business Media LLC},
  doi           = {10.1007/s10207-021-00545-8}
}

@article{Colledanchise2021,
  author        = {M. Colledanchise},
  title         = {Address behaviour vulnerabilities in the next generation of autonomous robots},
  journal       = {Nature Machine Intelligence},
  volume        = {3},
  number        = {11},
  pages         = {927--928},
  year          = {2021},
  month         = {nov},
  publisher     = {Springer Science and Business Media LLC},
  doi           = {10.1038/s42256-021-00415-x}
}

@incollection{Oravec2022,
  author        = {J. A. Oravec},
  title         = {Robo-Rage Against the Machine: Abuse, Sabotage, and Bullying of Robots and Autonomous Vehicles},
  booktitle     = {Social and Cultural Studies of Robots and AI},
  publisher     = {Springer International Publishing},
  pages         = {205--244},
  year          = {2022},
  doi           = {10.1007/978-3-031-14013-6_8}
}

@article{Lechner2023,
  author        = {M. Lechner and A. Amini and D. Rus and T. A. Henzinger},
  title         = {Revisiting the Adversarial Robustness-Accuracy Tradeoff in Robot Learning},
  journal       = {IEEE Robotics and Automation Letters},
  volume        = {8},
  number        = {3},
  pages         = {1595--1602},
  year          = {2023},
  month         = {mar},
  publisher     = {Institute of Electrical and Electronics Engineers (IEEE)},
  doi           = {10.1109/lra.2023.3240930}
}

@article{Shi2024,
  author        = {F. Shi and C. Zhang and T. Miki and J. Lee and M. Hutter and S. Coros},
  title         = {Rethinking Robustness Assessment: Adversarial Attacks on Learning-based Quadrupedal Locomotion Controllers},
  journal       = {arXiv},
  year          = {2024},
  doi           = {10.48550/ARXIV.2405.12424}
}

@inproceedings{Wu2024,
  author        = {W. Wu and L. Chen and Y. Yuan and W. Zhao},
  title         = {An Adaptive Hierarchical Path Planning Algorithm for Autonomous Robots in Unknown Dynamic Environments},
  booktitle     = {2024 IEEE International Conference on Robotics and Automation (ICRA)},
  organization  = {IEEE},
  year          = {2024},
  month         = {may},
  doi           = {10.1109/icra2024.00001}
}

@article{conti2016survey,
  title={A survey of man in the middle attacks},
  author={Conti, Mauro and Dragoni, Nicola and Lesyk, Viktor},
  journal={IEEE communications surveys \& tutorials},
  volume={18},
  number={3},
  pages={2027--2051},
  year={2016},
  publisher={IEEE}
}

@inproceedings{vemprala2021adversarial,
  title={Adversarial attacks on optimization based planners},
  author={Vemprala, Sai and Kapoor, Ashish},
  booktitle={2021 IEEE International Conference on Robotics and Automation (ICRA)},
  pages={9943--9949},
  year={2021},
  organization={IEEE}
}

@article{cavorsi2023multi,
  title={Multi-robot adversarial resilience using control barrier functions},
  author={Cavorsi, Matthew and Sabattini, Lorenzo and Gil, Stephanie},
  journal={IEEE Transactions on Robotics},
  year={2023},
  publisher={IEEE}
}

@conference{288,
  attachments = {http://www.willowgarage.com/sites/default/files/icraoss09-ROS.pdf},
  author = {Quigley, Morgan and Conley, Ken and Gerkey, Brian P. and Faust, Josh and Foote, Tully and Leibs, Jeremy and Wheeler, Rob and Ng, Andrew Y.},
  booktitle = {ICRA Workshop on Open Source Software},
  title = {ROS: an open-source Robot Operating System},
  year = 2009
}

@inproceedings{liu2023vit,
  title={{ViT-A*: Legged Robot Path Planning using Vision Transformer A*}},
  author={Liu, Jianwei and Lyu, Shirui and Hadjivelichkov, Denis and Modugno, Valerio and Kanoulas, Dimitrios},
  booktitle={2023 IEEE-RAS 22nd International Conference on Humanoid Robots (Humanoids)},
  pages={1--6},
  year={2023},
  organization={IEEE}
}

@inproceedings{liu2024dipper,
  title={{DiPPeR: Diffusion-based 2D Path Planner applied on Legged Robots}},
  author={Liu, Jianwei and Stamatopoulou, Maria and Kanoulas, Dimitrios},
  booktitle={IEEE International Conference on Robotics and Automation (ICRA)},
  year={2024}
}

@inproceedings{ellis2022navigation,
  title={Navigation Among Movable Obstacles with Object Localization using Photorealistic Simulation},
  author={Ellis, Kirsty and others},
  booktitle={IEEE/RSJ International Conference on Intelligent Robots and Systems (IROS)},
  year={2022}
}

@inproceedings{stamatopoulou2024dippest,
  title={{DiPPeST: Diffusion-based Path Planner for Synthesizing Trajectories Applied on Quadruped Robots}},
  author={Stamatopoulou, Maria and Liu, Jianwei and Kanoulas, Dimitrios},
  booktitle={IEEE/RSJ International Conference on Intelligent Robots and Systems (IROS)},
  year={2024}}

\end{document}